# Fingerprint recognition using standardized fingerprint model

Le Hoang Thai [1] and Ha Nhat Tam [2]

[1] Faculty of Information Technology, University of Science
Ho Chi Minh City, 70000, Viet Nam

[2] University of Science
Ho Chi Minh City, 70000, Viet Nam

**Abstract**
Fingerprint recognition is one of most popular and accuracy Biometric technologies. Nowadays, it is used in many real applications. However, recognizing fingerprints in poor quality images is still a very complex problem. In recent years, many algorithms, models… are given to improve the accuracy of recognition system. This paper discusses on the *standardized fingerprint model* which is used to synthesize the template of fingerprints. In this model, after pre-processing step, we find the transformation between templates, adjust parameters, synthesize fingerprint, and reduce noises. Then, we use the final fingerprint to match with others in FVC2004 fingerprint database (DB4) to show the capability of the model.

## 1. Introduction

Nowadays, fingerprint recognition is one of the most important biometric technologies based on fingerprint distinctiveness, persistence and ease of acquisition. Although there are many real applications using this technology, its problems are still not fully solved, especially in poor quality fingerprint images and when low-cost acquisition devices with a small area are adopted.

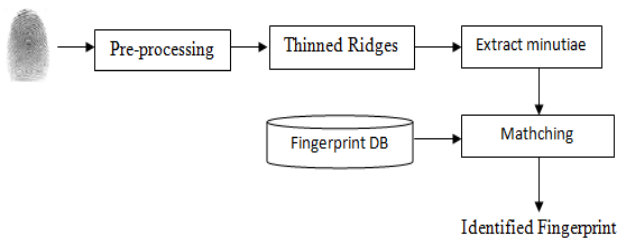

Fig.1 Fingerprint Recognition process

In fingerprint recognition process, the important step which affects on system accuracy is matching between template and query fingerprint. Many solutions are designed to increase this step's accuracy ([1], [2], [5], [6], [7], [9]). These matching algorithms may be classified into three types: minutiae-based approach, correlation-based approach and feature-based approach. However, as [9] analyzed, the score of these algorithms is not high (especially in case fingerprints are of the same finger but they are rotated or the intersection is too small). So, it's necessary to design a model to standardized fingerprint template in order to improve matching score.

In this paper, we propose a *standardized fingerprint model* to synthesize fingerprints which represents for all fingerprint templates stored in database when matching. The experimental results on DB4 (FVC2004 fingerprint database) show the capability of the model.

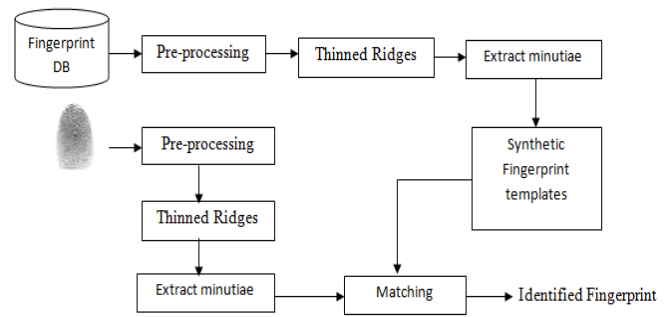

Fig.2 Fingerprint recognition using Standardized fingerprint model

## 2. A model of standardized fingerprint

### 2.1 Fingerprint features

A fingerprint is the reproduction of a fingertip epidermis, produced when a finger is pressed against a smooth surface. The most evident structural characteristic of a fingerprint is its pattern of interleaved ridges and valleys. Ridges and valleys often run parallel but they can bifurcate or terminate abruptly sometimes.

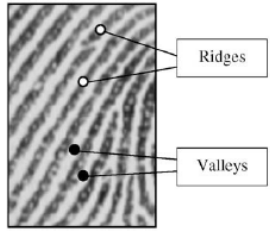

Fig.3 Ridges and valleys on a fingerprint image

The minutia, which is created when ridges and valleys bifurcate or terminate, is important feature for matching algorithms.

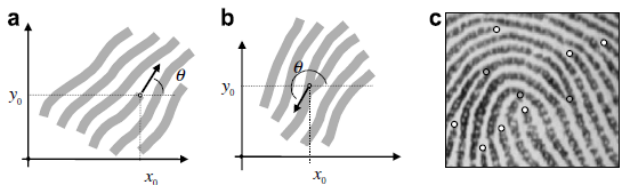

Fig.4 (a) A termination minutia (b) bifurcation minutia (c) termination (white) and bifurcation (gray) minutiae in a sample fingerprint

The fingerprint pattern contains one or more regions where the ridge lines create special shapes. These regions may be classified into three classes: *loop*, *delta*, and *whorl*. Many fingerprint matching algorithms pre-align fingerprint images based on a landmark or a center point which is called the *core* (Fig 5).

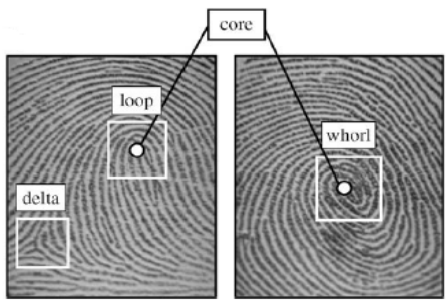

Fig.5 Special regions (white boxes) and core points (small circles) in fingerprint images

## 2.2 Standardized fingerprint model

From the given images of fingerprint, which are low quality or scaled or rotated together, we propose a model to create a new fingerprint image, which contains features (ridge line and minutia) of the original ones. The model includes the following steps: (1) Pre-processing fingerprint image: for each image, we recognize fingerprint area, thinned ridge lines and extract minutiae. (2) Finding and adjusting parameter sets: at first, choose a fingerprint which has largest fingerprint area as mean image. Then, we use Genetic Algorithms in [9] to find the transformation between mean image and others. (3) Synthesizing fingerprint: with the transformations in previous step, we re-calculate parameters' value (to get exact value for parameters), add supplement ridge lines and minutiae to mean fingerprint. (4) Post-processing: this step will help removing the noise of step 3.

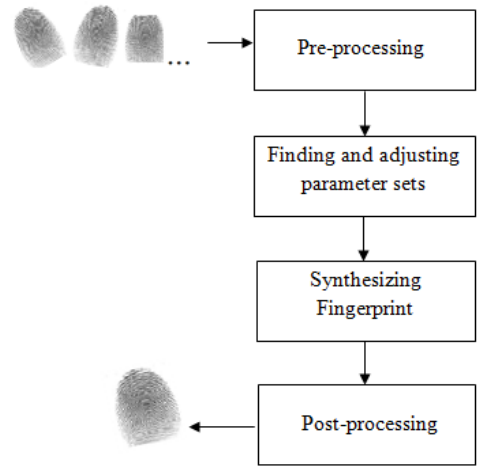

Fig.6 Synthesizing Fingerprint Model

*Pre-processing fingerprint:*

For each input image, we find fingerprint area and thin ridge line whose width is 1 pixel. P is a point on processed fingerprint image and pixel(P) is value of pixel at P:
- Pixel(P) = 1 if P belong to ridge
- Pixel(P) = 0 if P belong to valley

Each minutia, we get in this step, contains the x- and y-coordinates, the type (which is termination or bifurcation) and the angle between the tangent to the ridge line at the minutia position and the horizontal axis. Result of this step is a processed fingerprint called *Flist*

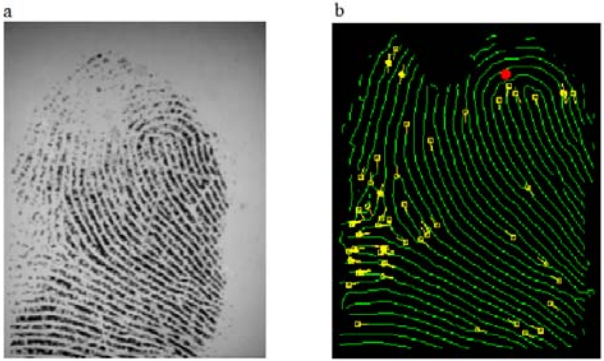

Fig.7 (a) Fingerprint image in database   (b) Fingerprint image after pre-processing step

*Finding and adjusting parameter set:*

Base on the result of pre-processing step, we use the Genetic Algorithm which is proposed by Tan and Bhanu in [9] to find the transformation between *meanF* (a fingerprint which has the largest fingerprint area as mean fingerprint) and others in *FList*. And then, we re-calculate the exact value of these parameters.

*Step 1*: Find parameter set: In [9], Tan and Bhanu proposed a transformation:

$$Y_i = F(X_i) = s.R.X_i + T \quad (1)$$

Where $s$ is the scale factor

$$R = \begin{bmatrix} \cos\theta & -\sin\theta \\ \sin\theta & \cos\theta \end{bmatrix}$$

$\theta$ : angle of rotation between two fingerprints
$T = [t_x, t_y]$ is the vector of translation.

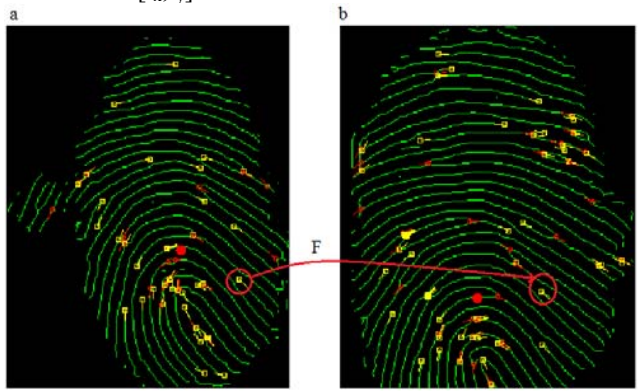

Fig.8 (a) fingerprint template (b) *meanF and the transformation* $F(s, \theta, t_x, t_y)$

Parameter set contains several parameters. Each parameter has form of $<s, \theta, t_x, t_y>$. To build parameter set, we perform:

**Input**: fingerprint template *FList*
**Output**: parameter set *ParamList*

1. *meanF* = fingerprint which has the largest fingerprint area
2. remove *meanF* from *FList*
3. For each *fData* in *FList*:
    a. *param* = Find the transformation between *meanF* and *fData*
    b. add *param* to *ParamList*

After finishing step 1, we perform the following tasks to re-calculate exact value of parameter in step 2: re-calculate exact value of parameter:

**Input**: *FList*, *ParamList*
**Output**: *ParamList* with real value of parameters
For each *fData* in *FList*:
1. Find 2 minutiae A, B in *fData* and 2 minutiae C, D in *meanF* in which A is corresponding to C and B is corresponding D.
2. Calculate the real value for parameters:
    a. s = sign(old value of s)[( distance between C and D) /( distance between A and B)]
    b. $\theta$ = sign(old value of $\theta$)[the angle between $\vec{AB}$ and $\vec{CD}$]
    c. $t_x$ = sign(old value of $t_x$) $|(x_A - x_C)|$
    d. $t_y$ = sign(old value of $t_y$) $|(y_A - y_C)|$
3. Update new value for corresponding parameter of *fData*

*Synthesizing fingerprint*

After re-calculating, new value of parameter set is used to add ridge lines and minutiae from the original fingerprint, which *meanF* does not have, to *meanF*. Synthesizing fingerprint contains three steps: (1) Add ridges from original fingerprint to *meanF* (2) Join supplement and original ridge lines (3) Add minutiae from original fingerprint to *meanF*

*Add ridge lines to meanF:* with each fingerprint in *FList*, we use correspondent parameter in *ParamList* to transform each of its pixels to *meanF's* space and put the pixel which doesn't have corresponding point in *meanF*. Finally, fill marked pixels to *meanF*

**Input**: *FList, ParamList*
**Output**: new *meanF*
For each $FList_k$ in *FList*
1. Using parameter k in *ParamList* to transform all pixel of $FList_k$ to *meanF* space and save to *PixelList*
2. For each pixel H in *PixelList*
    If *meanF* doesn't have pixel H' which d(H,H') < r and pixel(H') =1 then put H to *tempPixelList*
3. For each pixel H in *tempPixelList*
    Set Pixel(H)$_{meanF}$ = 1

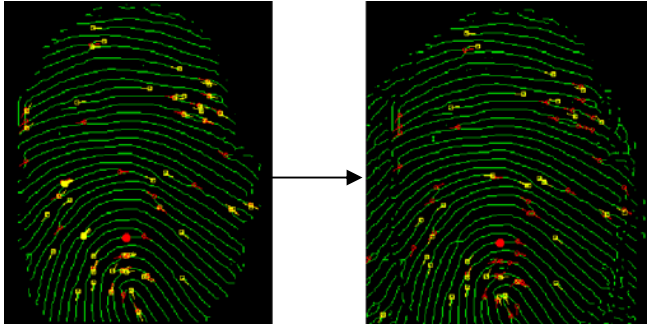

Fig.9 *meanF* before and after Synthesizing

*Join supplement and original ridge lines:* in previous step, we got a new *meanF* as above picture. However, the ridges were broken because algorithm in *Add ridges to meanF* step does not affect on the point which is the end of ridge line

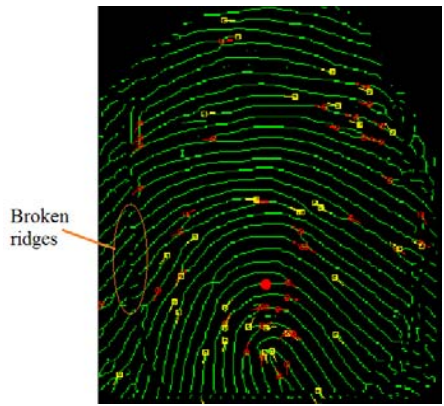

Fig.10 Some ridges were broken after Synthesizing

To solve this problem, we perform:

**Input**: *FList, ParamList, meanF*
**Output**: new *meanF*
1. For each *fData* in *FList*
   1.1. For each pixel K in *meanF*
       If K is the terminated point (pixel(K) =1) and exist K' in *fData* (pixel(K') =1) which correspondent to K then:
       - Find all pixels N (pixel(N)=1) connected to pixel K'
       - Transform these pixels to meanF's space
       - Put them to *linkedPixelList*
   1.2. Fill all pixels in *linkedPixelList* to meanF

To perform *Find all pixels N connected to pixel K'* task, use below algorithm:

**Input**: *FList, ParamList, meanF, K'*
**Output**: all pixels connected to pixels K'
1. Get all pixels, which is 8-neighbour of K' and pixels value is 1, save to *connectedList*
2. For each pixel M in *connectedList*
   a. Set *markedPixel* = K'
   b. Set startPixel = M
   c. Repeat follow tasks:
      - Get pixel L, which is 8-neighbour of M, pixel(M) =1 and M is different to *markedPixel*, save to *connectedList*
      - *markedPixel* = M
      - *startPixel* = L
      Until (can't find L or exist L' in *meanF* corresponding to L)

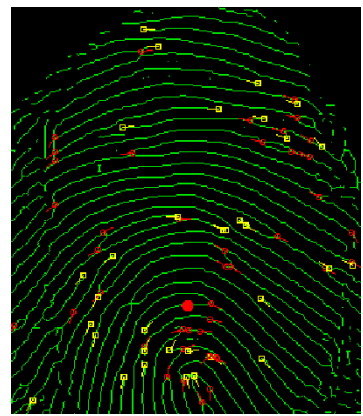

Fig.11 *meanF* after ridges connection step

*Post-Processing:*

Synthesizing fingerprint step creates a fingerprint image that contains all features of fingerprint templates. However, some minutiae of the original fingerprint are not correct on *meanF*. For example, M is termination minutia on fingerprint template but in *meanF*, it is not correct because of ridge line connection. In this step, we re-check all *meanF*'s minutiae and remove wrong minutiae.

**Input**: *meanF*
**Output**: *meanF* with *minutiaeList* which is removed wrong minutiae.
1. For each minutiae M in *minutiaeList* of *meanF*
   - If type of M is termination minutia and pixel(M) = 1 and M is termination point then M is marked
   - If type of M is bifurcation minutia and pixel(M) = 1 and M is not termination point then M is marked
2. Remove all un-marked minutiae from *minutiaeList*

## 3. Experiment result

3.1 Database

Database used for experiment is DB4 FVC2004. Several fingerprint images in this database are low quality. Size of each fingerprint images is 288x384 pixels, and its resolution is 500 dpi. FVC2004 DB4 has 800 fingerprints of 100 fingers (8 images for each finger). Fingerprint images are numbered from 1 to 100 followed by a another number (from 1 to 8) which mean that the image fingerprint is first to $8^{th}$ impression of certain finger (Fig.12).

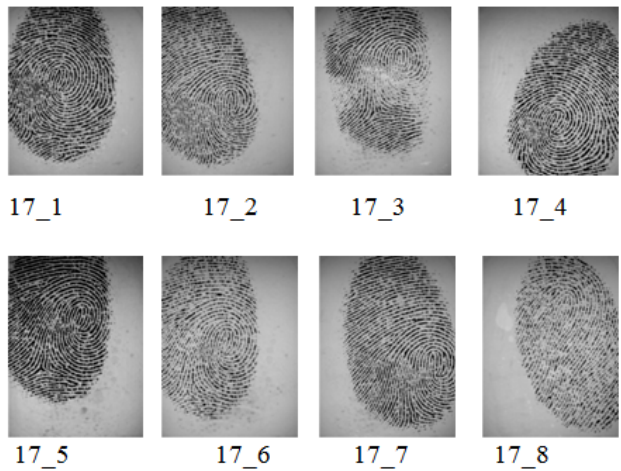

Fig.12 Sample of fingerprint in DB4

3.2 Estimation of data range model

The estimation of the data range is based on the experiments of the first 100 pairs of fingerprints in FVC2004 DB4. Base on the experiment results and data range in [9], the data range that we choose for these parameters are:

- $0.97 <= s <= 1.2$
- $-30º <= θ <= 30º$
- $-114 <= tx <= 152$
- $-128 <= ty <= 156$

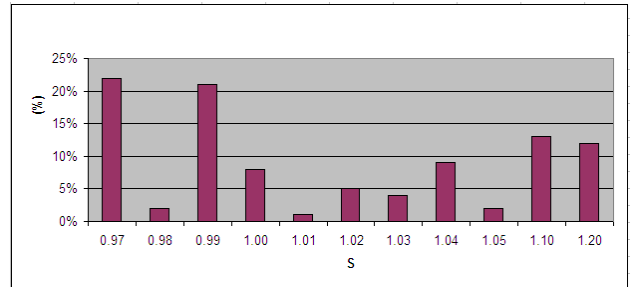

Fig.13 Data range of scale factor

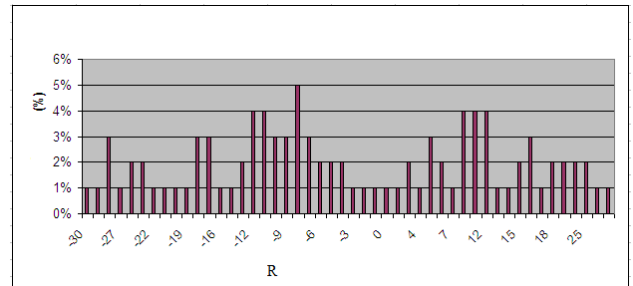

Fig.14 Data range of R

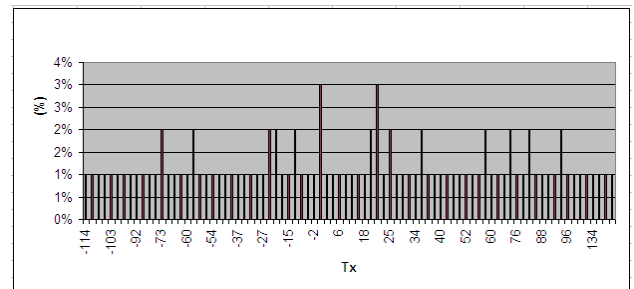

Fig.15 Data range of $T_x$

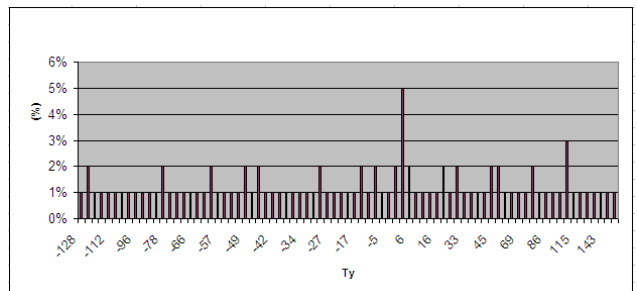

Fig.16 Data range of $T_y$

### 3.3 Estimation of data range model

Base on [9] and experiment results, threshold and value of parameter are chosen as below table:

Table 1: Parameter

| Parameter | Value | Parameter | Value |
|---|---|---|---|
| $T_d$ | 10 | $N_t$ | 15 |
| $P_m$ | 0.1 | $P_s$ | 0.8 |
| $N_p$ | 500 | r | 3 |

### 3.4 Result

The experiment is performed on Intel (R), CPU T3400 2.16 GHz with 1Gb RAM computer. A total of 800 synthesizings, 2800 matchings between consistent pairs are performed to estimate the distribution of genuine matching. We also perform 79200 matchings between inconsistent pairs to estimate the distribution of imposter matching, where for each matching we randomly select two fingerprints from FVC2004 DB4 that are the impressions of different fingers. The experiment results are compared to another results based on approach of Xiping Luo, 2000.

Table 2: Compare GAR and FAR of our approach and Xiping Luo, 2000

| | Our approach | | Xiping Luo, 2000 | |
|---|---|---|---|---|
| | GAR | FAR | GAR | FAR |
| 1 | 100% | 7.14% | 100% | 39% |
| 2 | 98.99% | 7.12% | 99% | 32% |
| 3 | 97.98% | 3.57% | 97% | 25% |
| 4 | 96% | 3.57% | 94% | 25% |
| 5 | 95% | 3.57% | 90% | 14.28% |

Table 3: Top 10 matchings of our approach and Xiping Luo

| # | Our approach | Xiping Luo, 2000 |
|---|---|---|
| 1 | 98.3% | 92.3% |
| 2 | 98.9% | 93.6% |
| 3 | 99.1% | 94.1% |
| 4 | 99.1% | 94.1% |
| 5 | 99.1% | 94.1% |
| 6 | 99.2% | 94.6% |
| 7 | 99.3% | 94.9% |
| 8 | 99.4% | 95.2% |
| 9 | 99.4% | 95.2% |
| 10 | 99.4% | 95.2% |

## 4. Conclusions

In this paper, we proposed a fingerprint-matching approach, which is based on *standardized fingerprint model* to synthesize fingerprint from original templates. From the fingerprint templates of finger in the database, we chose one as mean images and use Genetic Algorithms in [9] to find the transformation among them. Then, these transformations is used to synthesize fingerprints (add ridges and minutiae from original template to mean fingerprint). Finally, we perform matching between mean fingerprint and other templates (FVC2004 DB4 database, which has poor-quality fingerprints) to show the capability of the model.

**Dr Le Hoang Thai** received B.S degree and M.S degree in Computer Science from Hanoi University of Technology, Viet Nam, in 1995 and 1997. He received Ph.D. degree in Computer Science from HoChiMinh University of Natural Sciences, Vietnam, in 2004. Since 1999, he has been a lecture at Faculty of Information Technology, HoChiMinh University of Natural Sciences, Vietnam. He research interests include soft computing pattern recognition, image processing, biometric and computer vision. Dr. Le Hoang Thai is co-author over twenty papers in international journals and international conferences.

**Ha Nhat Tam** received B.S degree in Computer Science from HoChiMinh University of Natural Sciences, Vietnam in 2005. He is currently pursuing M.S degree in Computer Science HoChiMinh University of Natural Sciences.